\documentclass[a4paper,twoside]{article}

\usepackage{epsfig}
\usepackage{subcaption}
\usepackage{calc}
\usepackage{amssymb}
\usepackage{amstext}
\usepackage{amsmath}
\usepackage{amsthm}
\usepackage{multicol}
\usepackage{pslatex}
\usepackage{apalike}
\usepackage{algorithm2e}
\usepackage[bottom]{footmisc}
\usepackage{SCITEPRESS}     

\begin{document}

\title{A multifractal-based masked auto-encoder: an application to medical images}

\author{\authorname{Joao Batista Florindo\sup{1}\orcidAuthor{0000-0002-0071-0227} and Viviane de Moura\sup{1}}
\affiliation{\sup{1}Institute of Mathematics, Statistics and Scientific Computing - University of Campinas, Rua S\'{e}rgio Buarque de Holanda, 651, Cidade Universit\'{a}ria "Zeferino Vaz" - Distr. Bar\~{a}o Geraldo, CEP 13083-859, Campinas, SP, Brasil}
\email{florindo@unicamp.br,v215939@dac.unicamp.br}
}


\abstract{Masked autoencoders (MAE) have shown great promise in medical image classification. However, the random masking strategy employed by traditional MAEs may overlook critical areas in medical images, where even subtle changes can indicate disease. To address this limitation, we propose a novel approach that utilizes a multifractal measure (Renyi entropy) to optimize the masking strategy. Our method, termed Multifractal-Optimized Masked Autoencoder (MO-MAE), employs a multifractal analysis to identify regions of high complexity and information content. By focusing the masking process on these areas, MO-MAE ensures that the model learns to reconstruct the most diagnostically relevant features. This approach is particularly beneficial for medical imaging, where fine-grained inspection of tissue structures is crucial for accurate diagnosis. We evaluate MO-MAE on several medical datasets covering various diseases, including MedMNIST and COVID-CT. Our results demonstrate that MO-MAE achieves promising performance, surpassing  other basiline and state-of-the-art models. The proposed method also adds minimum computational overhead as the computation of the proposed measure is straightforward. Our findings suggest that the multifractal-optimized masking strategy enhances the model's ability to capture and reconstruct complex tissue structures, leading to more accurate and efficient medical image representation. The proposed MO-MAE framework offers a promising direction for improving the accuracy and efficiency of deep learning models in medical image analysis, potentially advancing the field of computer-aided diagnosis.}

\onecolumn \maketitle \normalsize \setcounter{footnote}{0} \vfill

\section{\uppercase{Introduction}}

Self-supervised learning (SSL) has emerged as a powerful paradigm in modern deep learning, offering a promising approach to overcome the limitations of traditional supervised and unsupervised methods \cite{doersch2017multi}. The approach has gained significant traction in recent years, particularly in domains such as computer vision and applications as in computer-aided diagnostics \cite{krishnan2022self}. Masked Autoencoder (MAE) \cite{he2022masked} is an example of well-succeeded SSL method. MAEs work by reconstructing images from partially masked inputs, encouraging the model to learn meaningful representations by aggregating contextual information.

However, the random masking strategy employed by traditional MAEs may not be optimal for medical images, where subtle changes in specific regions can be crucial for accurate diagnosis. In medical imaging, certain areas often contain more diagnostically relevant information than others. For instance, in chest X-rays, the lung fields typically hold more critical information for detecting respiratory diseases compared to the surrounding areas. To address this limitation, researchers have explored approaches to optimize the masking strategy of MAEs, also in applications to medical images \cite{mao2024medical}. 

One promising direction to quantify the relevance of image regions and consequently guide MAE masking process, and that has not yet been explored in the literature for this purpose, is multifractal analysis. This has been successfully applied in image processing and pattern recognition applications, e.g., in texture analysis and classification \cite{florindo2023randomized}. Multifractal analysis provides a framework for describing the complexity and heterogeneity of images across different scales, making it particularly suitable for capturing the intricate structures often present in real-world images. One of the most effective and straightforward techniques for multifractal analysis in digital images is Renyi entropy \cite{renyi1961measures}. This is a generalization of Shannon entropy and has been used in image processing, for example in texture recognition \cite{florindo2023renyi}. Its ability to characterize the information content of images at different scales makes it a potential candidate for optimizing masking strategies in MAEs for medical imaging. 

Building upon these foundations, this paper introduces a novel approach that combines the strengths of MAEs with multifractal analysis to enhance medical image classification. By utilizing Renyi entropy as a multifractal measure to guide the masking process, our proposed Multifractal-Optimized Masked Autoencoder (MO-MAE) aims to focus on regions of high complexity and information content, ensuring that the model learns to reconstruct the most diagnostically relevant features. Our main contributions and innovations are as follows:
\begin{itemize}
	\item We develop a multifractal-based masking strategy for MAE, improving results on medical image classification in the literature; our approach can also be easily extended to other application domains, in general tasks related to image classification.
	\item Up to our knowledge, this is the first time that multifractal analysis (and Renyi entropy in particular) is associated with masked auto-encoders in the literature.
	\item Being even more general, this is the first time that a physics-based complexity measure is explored in the MAE masking process, as other masking strategies usually rely on learnable procedures.
	\item We assess the proposed methodology on the well-established benchmarks of medical images MedMNIST \cite{yang2023medmnist} as well as on the real-world task of predicting COVID cases - dataset COVID-CT \cite{yang2020covid}. Extensive evaluations and comparison with results recently published in the literature are performed over those databases to confirm the potential of our proposal.
\end{itemize}

The proposed MO-MAE model outperforms other literature approaches in most scenarios both in the benchmark datasets and on the COVID-CT problem. Overall, our results suggest that using multifractal analysis to guide the masking strategy in the MAE framework is a promising direction and can be further explored, including applications to other domains outside the medical field or even other tasks, such as segmentation, for instance.

\section{\uppercase{Related works}}

Masked Autoencoders (MAE) have emerged as a promising paradigm for self-supervised learning in computer vision, achieving state-of-the-art performance across various benchmark datasets \cite{he2022masked}. MAEs have also been investigated in medical applications, particularly in image analysis and classification tasks. For example, in electrocardiography analysis, MAE-based self-supervised learning has shown promise in improving model performance for detecting left ventricular systolic dysfunction, even with limited training data \cite{sawano2024applying}. An overview on this topic can be found in \cite{krishnan2022self}.

Improvements over the original MAE architecture have also been explored. Several of them have focused on more elaborate masking strategies. That is the case of \cite{bandara2023adamae}, where an adaptive masking is proposed, using an auxiliary network that samples visible tokens based on the semantic context. Another one is \cite{li2022semmae}, where the authors propose a semantic-guided masking strategy. This is achieved by encouraging the neural network to learn various information from intra-part patterns to inter-part relations. A study specifically focused on medical images is described in \cite{mao2024medical}, where the authors propose the use of attention maps obtained by a supervised procedure to conduct the masking process. Theoretical studies on the role of masking in MAEs have also been presented, as in \cite{zhang2022mask}. Our proposal goes in another direction here as we focus on the use of a complexity measure to guide the masking process. Among the advantages of such approach, we can mention the interpretability of the masking criterion and the fact that our model does not require the learning of extra parameters in the pre-training stage or any other external training algorithm.

Fractal and multifractal theory have also been explored in the literature of image analysis, especially in medical images. In \cite{ding2023fractal}, a fractal graph convolutional neural network is proposed for computer-aided diagnosis using histopathological images. In \cite{swapnarekha2024deep}, a review of fractal-based image analysis with pattern recognition is presented. The authors in \cite{motwani2024fractal} investigate the correlation between the fractal dimension computed on CBCT scans of edentulous patients on implant site with the bone density determined by Misch's classification. Renyi entropy, particularly, was explored for image classification, for example, in \cite{florindo2023renyi}, where it was employed to analyze representations of deep convolutional neural networks. The use of multifractal analysis to guide MAE masking strategy is a novelty of our study.

\section{\uppercase{Proposed Method}}

\subsection{Overall Model}\label{sec:overall}

Despite the effectiveness of masked autoencoders in the literature, space for improvements still exist. One of such possibilities concern the mask selection step. Although the well-established random masking approach is straightforward, it does not take into account particularities of each image, for example, regions with more or less relevant information. In this context, here we present MO-MAE, a novel approach to MAE masking using multifractal analysis. Multifractal theory was originally developed to analyze complex physical systems, but has also demonstrated potential in image analysis \cite{florindo2023randomized}, particularly quantifying the \textit{complexity} of image regions and, as a consequence, its importance for the global image representation.
\begin{figure*}
	\centering
	\includegraphics[width=\textwidth]{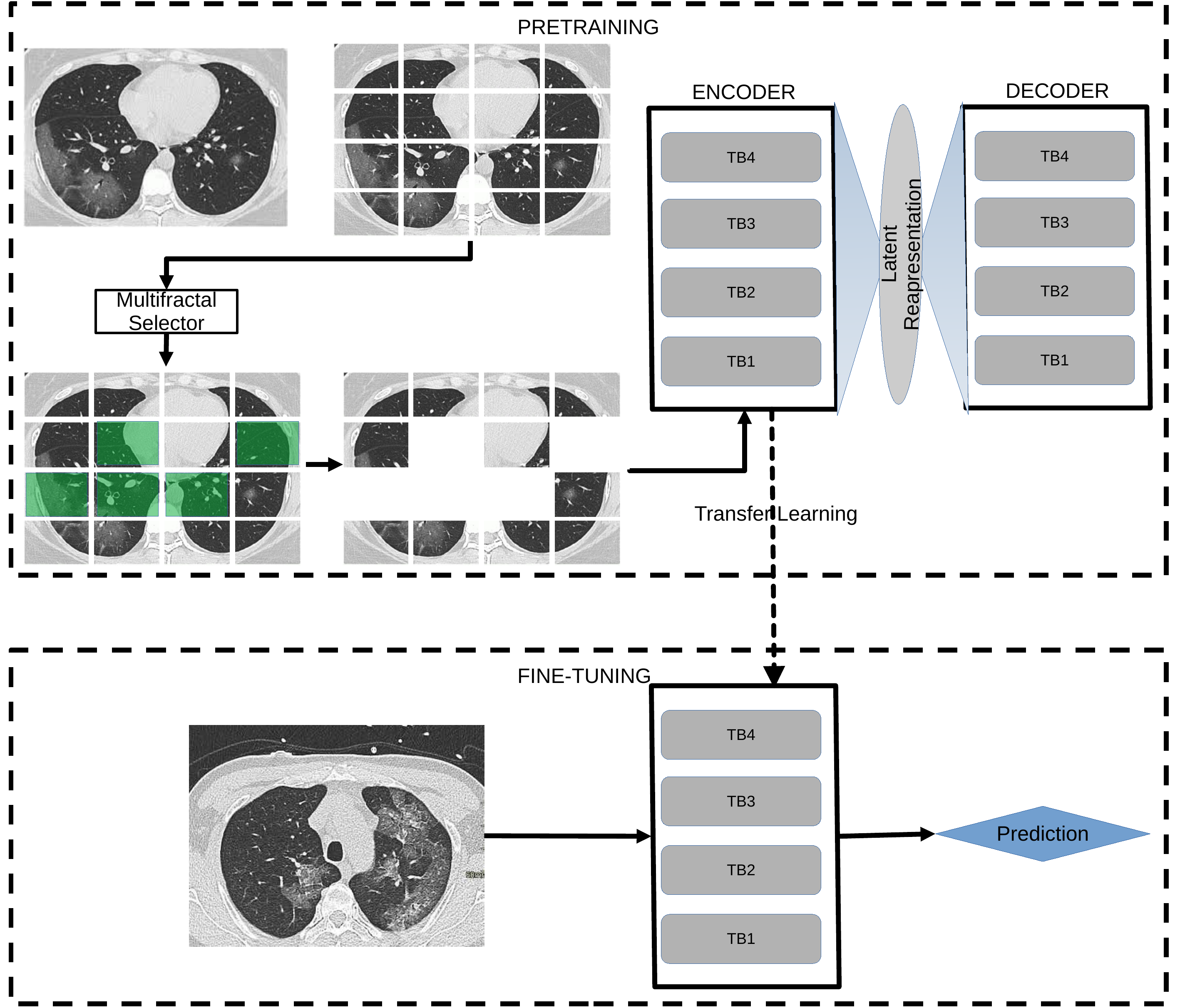}
	\caption{Proposed method. In the pretraining stage (upper block) we start by partitioning the image into rectangular patches (the number of patches here is only for illustrative purposes). Therefore we apply the multifractal selector module to identify patches with sufficiently relevant information. This is used as input to a ViT encoder, comprising 4 Transformer Blocks (TB). A mirrored architecture is used for decoding. The pretrained encoder is used in the fine-tuning step (lower block), to provide the deep latent representation of the input image and provide the desired prediction.}
	\label{fig:method}
\end{figure*}

Figure \ref{fig:method} provides a general schematic overview of the proposed methodology. As usual in self-supervised frameworks, the architecture is divided at high level into two tasks: the pretraining (upstream task) and fine-tuning (downstream). The overall model comprises the following major sequential steps:
\begin{enumerate}
	\item \textbf{Patching}: The image is partitioned into a collection of rectangular patches. The number and size of patches are hyperparameters to be determined by the user.
	\item \textbf{Multifractal analysis}: The multifractal spectrum is computed over each patch (more details on that in Section \ref{sec:multifractal}).
	\item \textbf{Masking}: Based on the multifractal spectrum, we select those patches with large amount of useful information. The percentage of selected patches is a hyperparameter.
	\item \textbf{Encoder/decoder pretraining}: The selected patches are used as input to an encoder module, which is a Vision Transformer (ViT). This is responsible for providing a latent representation of the input with reduced dimensionality. The output of the encoder is the input of another ViT, which plays the role of decoder. Both encoder and decoder are jointly trained with the objective of reconstructing the original image from the patches selected by the multifractal spectrum. The loss function measures the discrepancy between the original and the reconstructed images, as in \cite{he2022masked}.
	\item \textbf{Prediction (fine-tuning)}: Finally, the model receives the images of the target task (e.g., the diagnostic of a disease), previously labeled by a specialist or any other exogenous mechanism (e.g., a genetic test). This is processed by the encoder pretrained on the reconstruction task and this encoder is fine-tuned over the new labeled images. The final model is ready to be used on the test set and in the real-world application.
\end{enumerate}

\subsection{Multifractal Analysis}\label{sec:multifractal}

Our main novelty lies in the pretraining stage, in particular, in the multifractal selector, responsible for defining the patches that will be used as input to the reconstruction task. In \cite{falconer2013fractal}, two types of spectra are defined for multifractal analysis: the singularity and coarse spectra. For image analysis, given the limitation of the multiscale analysis imposed by the underlying resolution, the first one is more usual in general. Theoretical details can be found, for instance, in \cite{falconer2013fractal}, but in computational terms, we employ the partition function method \cite{salat2017multifractal}. Assuming a single-channel image $I:\mathbb{Z}^2\rightarrow\mathbb{Z}$, the codomain is partitioned with even spacing $s$, giving rise to
\[ y_j(s) = \sum_{i=(j-1)s+1}^{js}\mathbb{I}(I(\cdot)=i), \qquad 1\leq j\leq N_s=\lfloor G/s \rfloor, \]
where $\mathbb{I}$ is the indicator function and $G$ is the number of pixel intensity levels (default 255). From that, we estimate the probability distribution
\[ p_j(s) = \frac{y_j(s)}{\sum_{k=1}^{N_s} y_k(s)}. \]
The $q$-th moment partition function is therefore defined by
\[ Z_s^q = \sum_{j=1}^{N_s}[p_j(s)]^q, \]
which in a multifractal regime should obey the following power-law rule:
\[ Z_s^q \sim s^{\tau(q)}. \]
$\tau(q)$ is the scaling exponent function, also known as the multifractal spectrum of $I$. This also gives rise to an associated entropy, which is \textit{Renyi entropy}, defined by
\[ R_s^q = \frac{1}{1-q}\log(Z_s^q). \] 
The case $q=1$ is defined as being equivalent to the well-known Shannon entropy.

Here we divide the input image $I:\mathbb{Z}^{M\times N}\rightarrow \mathbb{Z}$ into $N'\times N'$ non-overlapping patches. The number of patches is $n_P = \lfloor N/N'\rfloor \times \lfloor M/N'\rfloor$. For the $k^{th}$ patch $P_k$ we compute the Renyi entropy $R_s^q(P_k)$. The patches are sorted in descendant order according to the entropy. Formally, let $\mathcal{P} = \left( P_k \right)_{k=1}^{n_P}$ be a sequence of patches such that $R_s^q(P_{k+1})\geq R_s^q(P_k)$. Provided the mask ratio $r\in[0,1]$, the number of selected patches is $n_S = (1-r)n_P$ and the patches are obtained from the subsequence 
\[\mathcal{P}' = \left( P_{n_k} \right)_{n_k=\{1,2,3,\cdots,n_S\}}.\]

The set of selected patches $\mathcal{P}'$ is finally introduced as input to the encoder in the pre-training state and all the remaining steps follows as described in Section \ref{sec:overall}. We ensure in this way that those patches with high complexity, as measured by Renyi entropy, are selected for the reconstruction. These also correspond to the richest regions on the image, and consequently those parts whose reconstruction is more challenging. By forcing the pretraining encoder to solve such difficult task, we gather more robust and richer features in the latent representation, which naturally will lead to more effectiveness in the target task, image classification in our case.


\section{\uppercase{Experimental Setup}}

For the implementation of the MAE algorithm, we adopted a patch size of $16\times 16$, 4 layers both in the encoder and decoder ViT, 200 epochs in the pretraining stage and 100 epochs for fine-tuning. The number of layers and pretraining epochs are considerably smaller than the original model, which used 12 layers and 2000 epochs, respectively. We observed that enlarging the backbone did not correspond to any significant improvement for our purposes. For the remaining hyperparameters we adopted default values, using AdamW as the optimizer. In the pretraining, we used a base learning rate of 1.5e-4, weight decay 0.05, batch size 4096 (MedMNIST) or 128 (COVID-CT), and mask ratio 0.75. In the fine-tuning, we used a base learning rate of 1e-3, weight decay 0.5, and batch size 128. The experiments were carried out on Google Colab environment with an Nvidia T4 GPU.

The performance of the proposed methodology was assessed on the collection of medical image datasets MedMNIST-V2 \cite{yang2023medmnist} and the COVID-CT database \cite{yang2020covid}. MedMNIST is a family of datasets, including both 2D and 3D biomedical images especially designed and preprocessed for benchmark. Here we use the 2D collection, which comprises a total of 708,000 labeled images, each one with size $224\times 224$. Those images cover a wide range of medical modalities (pathology, X-ray, dermatoscopy, retinal OCT, abdominal CT, breast ultrasound, etc.) and predictive tasks (binary/multi-class, ordinal regression, and multi-label). COVID-CT, on the other hand, consists of 349 COVID-19 CT and 397 Non-COVID-19 CT images. Those images were resized to 224 × 224. The dataset was split into a training, a validation, and a test set, by patient, with a ratio of 60\%, 15\%, and 25\%, respectively. 

As comparative metrics, we adopt accuracy (ACC), which is defined as the ratio of images correctly classified, area under the precision/recall curve (AUC), and F1 score (harmonic mean of precision and recall).
	
\section{\uppercase{Results and Discussion}}

\subsection{MedMNIST}

Table \ref{tab:ablation} presents the results of an ablation study, where we compare the model with and without the multifractal MAE module on MedMNIST datasets. We observe a general increase both in terms of accuracy and AUC. This is even more evident in the most challenging data, as in RetinaMNIST and BreastMNIST, but the superiority is consistent across all datasets.
\begin{table*}[!htpb]
	\centering
	\caption{Ablation experiment on MedMNIST datasets. The original MAE with classical masking strategy is compared with the multifractal-guided approach proposed here}
	\label{tab:ablation}	
	\scalebox{.8}{
		\begin{tabular}{|l||cc||cc|}
			\hline
			Dataset & \multicolumn{2}{c||}{Original} & \multicolumn{2}{c||}{MO-MAE (Proposed)}\\
			& AUC & ACC & AUC & ACC \\
			\hline
			PathMNIST & 0.996 & 0.948 & 0.997 & 0.953\\
			DermaMNIST & 0.922 & 0.806 & 0.959 & 0.810\\
			OCTMNIST & 0.992 & 0.917 & 0.993 & 0.917\\
			PneumoniaMNIST & 0.936 & 0.910 & 0.988 & 0.909\\
			RetinaMNIST & 0.734 & 0.497 & 0.822 & 0.588\\
			BreastMNIST & 0.855 & 0.885 & 0.920 & 0.872\\
			BloodMNIST & 0.999 & 0.989 & 1.000 & 0.988\\
			TissueMNIST & 0.930 & 0.709 & 0.944 & 0.720\\
			OrganAMNIST & 0.998 & 0.966 & 0.999 & 0.959\\
			OrganCMNIST & 0.995 & 0.941 & 0.998 & 0.939\\
			OrganSMNIST & 0.982 & 0.834 & 0.983 & 0.831\\
			\hline
			Average & 0.940$\pm$0.082 & 0.855$\pm$0.144 & \textbf{0.964}$\pm$0.054 & \textbf{0.862}$\pm$0.120\\
			\hline
	\end{tabular}}
\end{table*}

Another important investigation concerns the role of $q$ hyperparameter in the multifractal spectrum patch selection. This experiment is summarized by Table \ref{tab:q}. A classical intuition in multifractal theory relates $q$ with the role of a ``magnifying glass'': larger values of $q$ correspond to coarser scales of analysis. Here we see that $q=2$ is in general a compromise between short and long-range fractality observed over the patch. Based on that, this was our choice for the remaining experiments.
\begin{table*}[!htpb]
	\centering
	\caption{Evaluation of hyperparameter $q$ on MedMNIST datasets.}
	\label{tab:q}	
	\scalebox{.8}{
		\begin{tabular}{|l||cc||cc||cc|}
			\hline
			Dataset & \multicolumn{2}{c||}{$q=1$} & \multicolumn{2}{c||}{$q=2$} & \multicolumn{2}{c|}{$q=10$}\\
			& AUC & ACC & AUC & ACC & AUC & ACC\\
			\hline
			PathMNIST & 0.994 & 0.930 & 0.997 & 0.953 & 0.996 & 0.948\\
			DermaMNIST & 0.931 & 0.762 & 0.959 & 0.810 & 0.929 & 0.799\\
			OCTMNIST & 0.971 & 0.789 & 0.993 & 0.917 & 0.992 & 0.917\\
			PneumoniaMNIST & 0.978 & 0.929 & 0.988 & 0.909 & 0.955 & 0.909\\
			RetinaMNIST & 0.715 & 0.502 & 0.822 & 0.588 & 0.731 & 0.492\\
			BreastMNIST & 0.898 & 0.840 & 0.920 & 0.872 & 0.857 & 0.891\\
			BloodMNIST & 0.998 & 0.963 & 1.000 & 0.988 & 0.999 & 0.989\\
			TissueMNIST & 0.935 & 0.695 & 0.944 & 0.720 & 0.930 & 0.710\\
			OrganAMNIST & 0.998 & 0.948 & 0.999 & 0.959 & 0.999 & 0.967\\
			OrganCMNIST & 0.996 & 0.931 & 0.998 & 0.939 & 0.997 & 0.942\\
			OrganSMNIST & 0.982 & 0.821 & 0.983 & 0.831 & 0.983 & 0.836\\
			\hline
			Average & 0.949$\pm$0.081 & 0.828$\pm$0.139 & \textbf{0.964}$\pm$0.054 & \textbf{0.862}$\pm$0.120 & 0.942$\pm$0.083 & 0.854$\pm$0.145\\
			\hline
	\end{tabular}}
\end{table*}

\begin{figure*}[!htpb]
	\centering
	\begin{tabular}{c}
		\begin{tabular}{ccc}
			PathMNIST & DermaMNIST & OctMNIST\\
			\includegraphics[width=.33\textwidth]{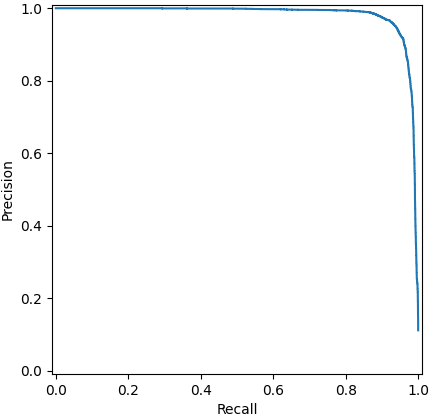}&
			\includegraphics[width=.33\textwidth]{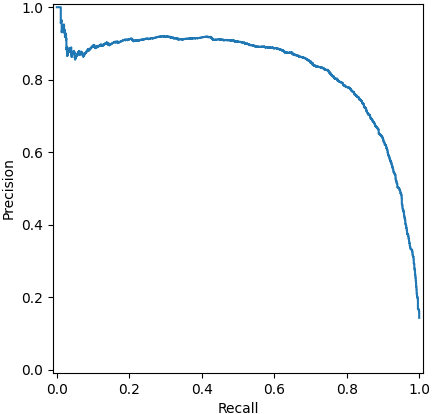}&
			\includegraphics[width=.33\textwidth]{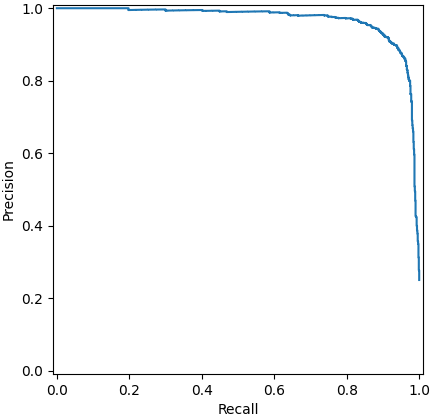}\\
		\end{tabular}\\
		\begin{tabular}{ccc}
			PneumoniaMNIST & RetinaMNIST & BreastMNIST\\		
			\includegraphics[width=.33\textwidth]{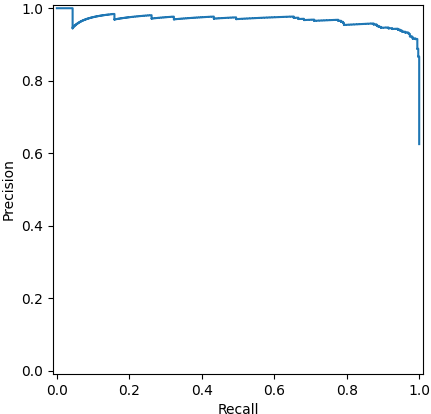}&
			\includegraphics[width=.33\textwidth]{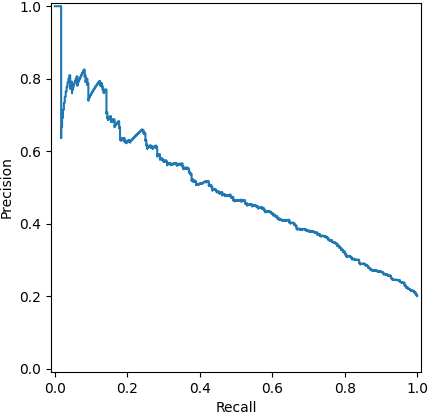}&
			\includegraphics[width=.33\textwidth]{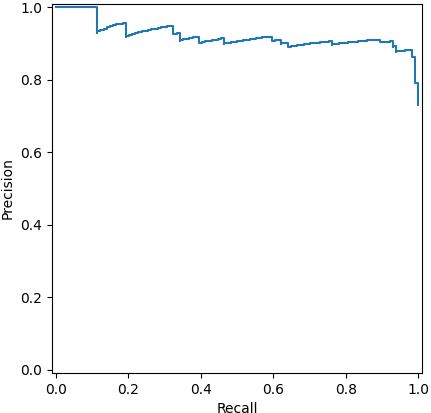}
		\end{tabular}\\
		\begin{tabular}{ccc}
			BloodMNIST & TissueMNIST & OrganAMNIST\\
			\includegraphics[width=.33\textwidth]{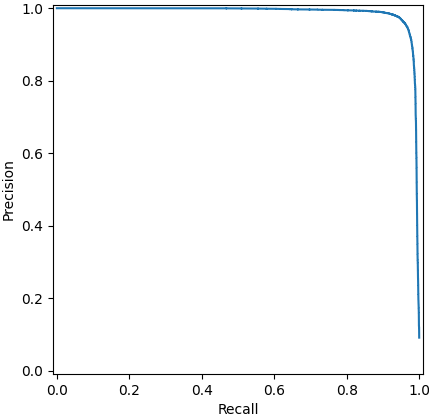}&
			\includegraphics[width=.33\textwidth]{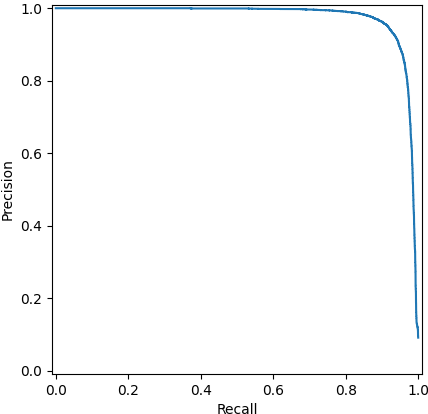}&
			\includegraphics[width=.33\textwidth]{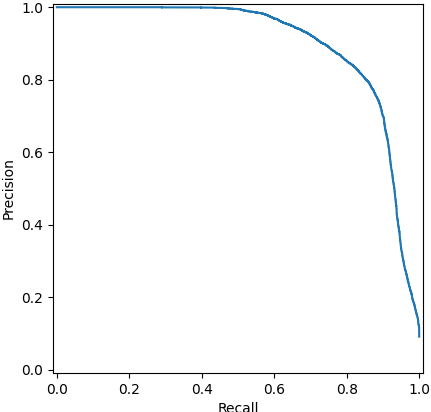}
		\end{tabular}\\
		\begin{tabular}{ccc}
			OrganCMNIST & OrganSMNIST & \\
			\includegraphics[width=.33\textwidth]{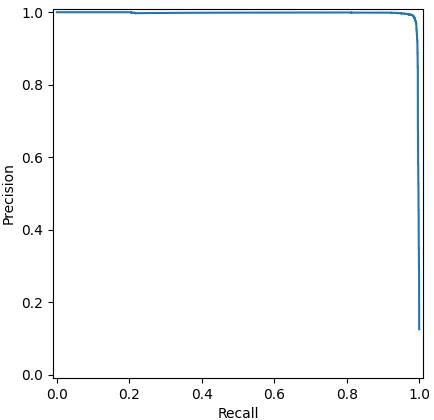}&
			\includegraphics[width=.33\textwidth]{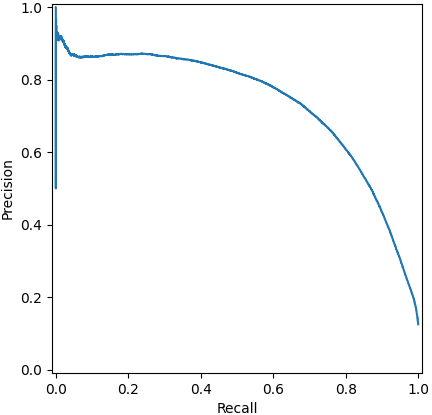} &
		\end{tabular}	
	\end{tabular}
	\caption{Precision/Recall curves for the proposed MO-MAE method on the MedMNIST datasets.}
	\label{fig:PR}
\end{figure*}

Figure \ref{fig:PR} depicts the precision/recall curves for the proposed method on the MedMNIST datasets. As already explicit in the AUC values of Table \ref{tab:q}, a more challenging scenario is noticeable for RetinaMNIST, followed by BreastMNIST, and TissueMNIST. RetinaMNIST and BreastMNIST have the smallest number of samples, which naturally makes any discriminative/predictive task more difficult for the machine learning algorithm. TissueMNIST, on the other hand, has the largest number of samples across all MedMNIST datasets, but the high variability of images within the same class and the number of categories also represent a significant challenge to automatic classification.
\addtolength{\tabcolsep}{-0.5em}
\addtolength{\tabcolsep}{+0.5em}

Table \ref{tab:literature} lists results recently published in the literature on MedMNIST datasets in comparison with the proposed approach. MO-MAE attains the highest AUC/ACC in most datasets. Here AUC is a more faithful metric considering possible imbalances in some of those datasets. And, with respect to AUC, the only exceptions where MO-MAE is the not the best method are PneumoniaMNIST, BreastMNIST, TissueMNIST, and OrganSMNIST. In all these cases, the highest AUC corresponds to MedVIT-S \cite{manzari2023medvit}. We should highlight, however, that this is a computationally intensive architecture from the state-of-the-art, combining deep Convolutional Neural Networks and Transformers. And even in these scenarios, our results are quite competitive. And it is also interesting to observe that MO-MAE outperformed MedVIT in most datasets, even considering the largest version MedVIT-L. Another point that is worth it to mention is the lack of competitiveness of fully automatic methods, such as Auto-sklearn, AutoKeras, and Google AutoML. Our results confirm that deep learning algorithms appropriately tuned for each particular task still is the best option in most practical situations.
\begin{table*}[!htpb]
	\centering
	\caption{Results for the proposed MO-MAE method on MedMNIST datasets compared with other methods in the literature. Literature results obtained from \cite{manzari2023medvit}.}
	\label{tab:literature}
	\begin{tabular}{c}
		\scalebox{.7}{
			\begin{tabular}{|l||cc||cc||cc||cc||cc||cc|}
				\hline
				Method & \multicolumn{2}{c||}{PathMNIST} & \multicolumn{2}{c||}{DermaMNIST} & \multicolumn{2}{c||}{OCTMNIST} & \multicolumn{2}{c||}{PneumoniaMNIST} & \multicolumn{2}{c||}{RetinaMNIST} & \multicolumn{2}{c|}{BreastMNIST}\\
				& AUC & ACC & AUC & ACC & AUC & ACC & AUC & ACC & AUC & ACC & AUC & ACC\\
				\hline
				ResNet-18 & 0.989 & 0.909 & 0.920 & 0.754 & 0.958 & 0.763 & 0.956 & 0.864 & 0.710 & 0.493 & 0.891 & 0.833\\
				ResNet-50 & 0.989 & 0.892 & 0.912 & 0.731 & 0.958 & 0.776 & 0.962 & 0.884 & 0.716 & 0.511 & 0.866 & 0.842\\
				Auto-sklearn & 0.934 & 0.716 & 0.902 & 0.719 & 0.887 & 0.601 & 0.942 & 0.855 & 0.690 & 0.515 & 0.836 & 0.803\\
				AutoKeras & 0.959 & 0.834 & 0.915 & 0.749 & 0.955 & 0.763 & 0.947 & 0.878 & 0.719 & 0.503 & 0.871 & 0.831\\
				Google AutoML & 0.944 & 0.728 & 0.914 & 0.768 & 0.963 & 0.771 & 0.991 & 0.946 & 0.750 & 0.531 & 0.919 & 0.861\\
				MedVIT-T & 0.994 & 0.938 & 0.914 & 0.768 & 0.961 & 0.767 & 0.993 & 0.949 & 0.752 & 0.534 & 0.934 & 0.896\\
				MedVIT-S & 0.993 & 0.942 & 0.937 & 0.780 & 0.960 & 0.782 & \textbf{0.995} & \textbf{0.961} & 0.773 & 0.561 & \textbf{0.938} & \textbf{0.897}\\
				MedVIT-L & 0.984 & \textbf{0.984} & 0.920 & 0.773 & 0.945 & 0.761 & 0.991 & 0.921 & 0.754 & 0.552 & 0.929 & 0.883\\
				\hline
				MO-MAE & \textbf{0.997} & 0.953 & \textbf{0.959} & \textbf{0.810} & \textbf{0.993} & \textbf{0.917} & 0.988 & 0.909 & \textbf{0.822} & \textbf{0.588} & 0.920 & 0.872\\
				\hline
		\end{tabular}}\\~\\~\\
		\scalebox{.7}{
			\begin{tabular}{|l||cc||cc||cc||cc||cc|}
				\hline
				Method & \multicolumn{2}{c||}{BloodMNIST} & \multicolumn{2}{c||}{TissueMNIST} & \multicolumn{2}{c||}{OrganAMNIST} & \multicolumn{2}{c||}{OrganCMNIST} & \multicolumn{2}{c|}{OrganSMNIST}\\
				& AUC & ACC & AUC & ACC & AUC & ACC & AUC & ACC & AUC & ACC\\
				\hline
				ResNet-18 & 0.998 & 0.963 & 0.933 & 0.681 & 0.998 & 0.951 & 0.994 & 0.920 & 0.974 & 0.778\\
				ResNet-50 & 0.997 & 0.950 & 0.932 & 0.680 & 0.998 & 0.947 & 0.993 & 0.911 & 0.975 & 0.785\\
				Auto-sklearn & 0.984 & 0.878 & 0.828 & 0.532 & 0.963 & 0.762 & 0.976 & 0.829 & 0.945 & 0.672\\
				AutoKeras & 0.998 & 0.961 & 0.941 & 0.703 & 0.994 & 0.905 & 0.990 & 0.879 & 0.974 & 0.813\\
				Google AutoML & 0.998 & 0.966 & 0.924 & 0.673 & 0.990 & 0.886 & 0.988 & 0.877 & 0.964 & 0.749\\
				MedVIT-T & 0.996 & 0.950 & 0.943 & 0.703 & 0.995 & 0.931 & 0.991 & 0.901 & 0.972 & 0.789\\
				MedVIT-S & 0.997 & 0.951 & \textbf{0.952} & \textbf{0.731} & 0.996 & 0.928 & 0.993 & 0.916 & \textbf{0.987} & 0.805\\
				MedVIT-L & 0.996 & 0.954 & 0.935 & 0.699 & 0.997 & 0.943 & 0.994 & 0.922 & 0.973 & 0.806\\
				\hline
				MO-MAE & \textbf{1.000} & \textbf{0.988} & 0.944 & 0.720 & \textbf{0.999} & \textbf{0.959} & \textbf{0.998} & \textbf{0.939} &  0.983 & \textbf{0.831}\\
				\hline
		\end{tabular}}
	\end{tabular}
\end{table*}

\subsection{COVID-CT}

\begin{figure}
	\centering
	\includegraphics[width=.5\textwidth]{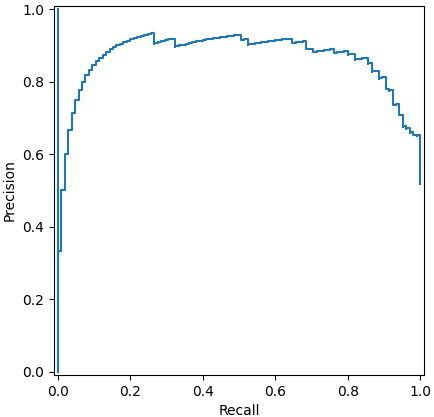}
	\caption{Precision/Recall curve for the proposed MO-MAE method on the COVID-CT dataset.}
	\label{fig:PR_covid}
\end{figure}

Figure \ref{fig:PR_covid} depicts the precision/recall curve for the proposed method on the COVID-CT database. Table \ref{tab:literature_covid} presents a comparison of our results on the COVID data with the literature. The curve is in line with the reported F1 score of 0.85 and follows a characteristic pattern where low recall also corresponds to low precision. This behavior is typically observed in nearly-balanced databases, which is the case of COVID-CT.

\begin{figure}
	\centering
	\includegraphics[width=.5\textwidth]{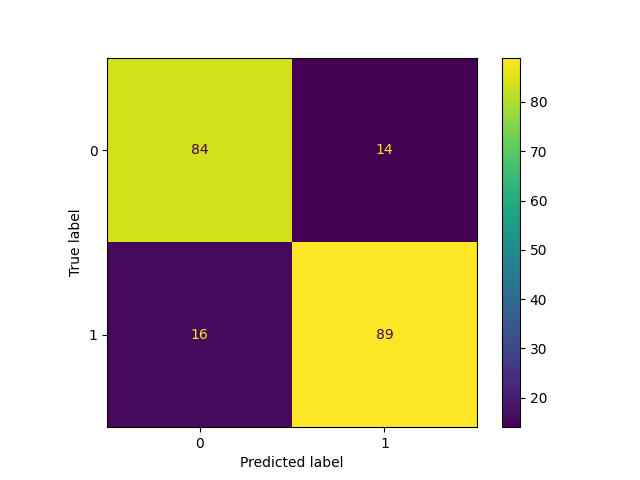}
	\caption{Confusion matrix for the proposed MO-MAE method on the COVID-CT dataset.}
	\label{fig:CM_covid}
\end{figure}

Figure \ref{fig:CM_covid} depicts the confusion matrix for MO-MAE on the COVID-CT dataset. We notice that our method performs well both with respect to the positive and negative classes.

Table \ref{tab:literature_covid} lists some results published in the literature for the COVID-CT database in comparison with MO-MAE in terms of accuracy and F1 score. Here we follow the protocol in \cite{abid2023multi}, which does not involve any pre-segmentation task, and consequently is more challenging that that explored in the original reference \cite{yang2020covid}. That is the reason why our method is not comparable with \cite{yang2020covid} but with \cite{abid2023multi}. MO-MAE achieves significant advantage, even over models with huge number of learnable parameters, such as DenseNet-169 and ResGANet-101.
\begin{table}[!htpb]
	\centering
	\caption{Results for MO-MAE on COVID-CT dataset compared with other methods in the literature. Literature results obtained from \cite{abid2023multi}.}
	\label{tab:literature_covid}
	\begin{tabular}{|lcc|}
		\hline
		Method & Accuracy & F1 Score\\
		\hline
		VGG-16 & 0.66 & 0.58\\
		ResNet-50 & 0.72 & 0.73\\
		DenseNet-169 & 0.80 & 0.79\\
		EfficientNet-b1 & 0.70 & 0.62\\
		CRNet & 0.72 & 0.76\\
		ShuffleNetV2 (1.5X) & 0.73 & 0.76\\
		SENet-50 & 0.76 & 0.77\\
		CBAM-50 & 0.78 & 0.80\\
		ResNeXt-50 & 0.72 & 0.75\\
		Res2Net-50 & 0.73 & 0.74\\
		ECANet-50 & 0.75 & 0.74\\
		SKNet-50 & 0.77 & 0.76\\
		ResGANet-101 (G=2) & 0.78 & 0.81\\
		\hline
		MO-MAE & \textbf{0.85} & \textbf{0.85}\\
		\hline
	\end{tabular}
\end{table}

Overall, the presented results suggest the potential of the proposed MO-MAE model as a powerful solution for medical image classification. The method outperformed several models with high computational burden in the literature and also demonstrated stability and robustness across different types of images and medical tasks.

\section{\uppercase{Conclusions}}

In this work, we proposed a new strategy for masking in masked auto-encoders. The masking process was guided by the multifractal spectrum computed over the image patches. Patches with the highest Renyi entropies were selected to compose the input to the pretraining task.

The efficiency of our proposal was assessed on benchmarks of medical images commonly used in the literature: MedMNIST collection of datasets and COVID-CT database. The obtained results are encouraging, demonstrating competitiveness with the state-of-the-art on medical image classification using deep learning. Particularly, our approach follows the self-supervised paradigm, which also makes it a naturally interesting solution in scenarios where the number of labeled images for training is limited. This is especially common in several areas of medicine.

Our approach can also be straightforwardly extended to other domains involving image classification and even related tasks, such as segmentation, for example. Future investigation on these possibilities are in progress. 

\section*{\uppercase{Acknowledgements}}

J. B. Florindo gratefully acknowledges the financial support of the S\~ao Paulo Research Foundation (FAPESP) (Grants \#2024/01245-1 and \#2020/09838-0) and from National Council for Scientific and Technological Development, Brazil (CNPq) (Grant \#306981/2022-0).


\end{document}